\DeclareUnicodeCharacter{FB01}{fi}
\documentclass{article}
\usepackage{spconf,amsmath,graphicx}
\usepackage{graphicx}
\usepackage{comment}
\usepackage{color}
\usepackage{amsmath}
\usepackage{amssymb}
\usepackage{subfigure}
\usepackage{booktabs} 
\usepackage{graphicx}
\usepackage[ruled]{algorithm2e}
\usepackage{multirow}
\newsavebox\CBox
\def\textBF#1{\sbox\CBox{#1}\resizebox{\wd\CBox}{\ht\CBox}{\textbf{#1}}}

\title{Lifelong Twin Generative Adversarial Networks}
\name{Fei Ye and Adrian G. Bors}
\address{Department of Computer Science, University of York, York YO10 5GH, UK}
\begin{document}
%
\maketitle
\begin{abstract}
In this paper, we propose a new continuously learning generative model, called the Lifelong Twin Generative Adversarial Networks (LT-GANs). LT-GANs learns a sequence of tasks from several databases and its architecture consists of three components: two identical generators, namely the Teacher and Assistant, and one Discriminator. In order to allow for the LT-GANs to learn new concepts without forgetting, we introduce a new lifelong training approach, namely Lifelong Adversarial Knowledge Distillation (LAKD), which encourages the Teacher and Assistant to alternately teach each other, while learning a new database. This training approach favours transferring knowledge from a more knowledgeable player to another player which knows less information about a previously given task. 
\end{abstract}


\vspace*{-0.2cm}
\section{Introduction}
\vspace*{-0.2cm}

An essential characteristic of human beings intelligence is that of being able to continually learn and acquire new skills and concepts from the world throughout their lifespan \cite{BranchSpecific}.  Neural networks trained on a sequence of tasks tend to focus on the latest learnt task and would perform poorly on any other learnt before. This phenomenon is called catastrophic forgetting \cite{Catastrophic}, and it is caused by the fact that network's parameters are overwritten each time when training on a new task.
A solution proposed to relieve catastrophic forgetting was to impose constraints while learning the tasks associated with a new database \cite{EWC}. This is achieved by including a regularization term in the objective function, penalizing the change in the network weights when learning a new task. Other solutions focused on dynamically increasing the number of neurons and network layers in order to be able to store novel information, \cite{Lifelong_expandable}. Most of these approaches require task labels, or the knowledge of the task boundaries, which is not always feasible. Moreover, these approaches are focused on the supervised learning setting while in this paper we address the more challenging problem of unsupervised learning, \cite{LifelongUnsupervisedVAE}. 

 Generative Adversarial Nets (GANs) \cite{GAN} can relieve catastrophic forgetting by being trained in a self-supervised fashion. Retraining with generative replay is achieved by building new training sets comprising of data generated by a GAN, which are added to a given real dataset from the current task \cite{MemoryReplayGAN}, or by preserving and freezing the model's parameters  after each dataset switch. However, applying these approaches in practical applications, that would favour a memory-efficient and end-to-end learning manner, is challenging. Moreover, existing GAN based lifelong learning models lack inference mechanisms and therefore can not capture complex structures behind data.

The proposed lifelong learning model, is named Lifelong Learning GANs (LT-GANs). LT-GANs relieves catastrophic forgetting by accumulating knowledge through a twin Teacher-Assistant network in the context of adversarial learning. This learning process favours transferring knowledge from a more knowledgeable player to its twin player during the lifelong learning, while the memory size is rather small and kept fixed. We further implement a Teacher-Student framework considering the LT-GANs as a Teacher network in order to learn data representations over time. 

This research study has the following contributions:
\begin{itemize}
\vspace*{-0.2cm}
	\item A new lifelong learning model, LT-GANs, which aims to learn a sequence of tasks from a set of databases. 
	\vspace*{-0.25cm}
\item We introduce the Lifelong Adversarial Knowledge Distillation (LAKD), an end-to-end training algorithm for accumulating knowledge across tasks.
\vspace*{-0.25cm}
\item  The LT-GANs model is extended to a Teacher-Student framework for learning data representations over time.
\end{itemize}

\begin{figure*}[htbp]
	\centering
	\vspace*{-0.5cm}
	\includegraphics[scale=0.7]{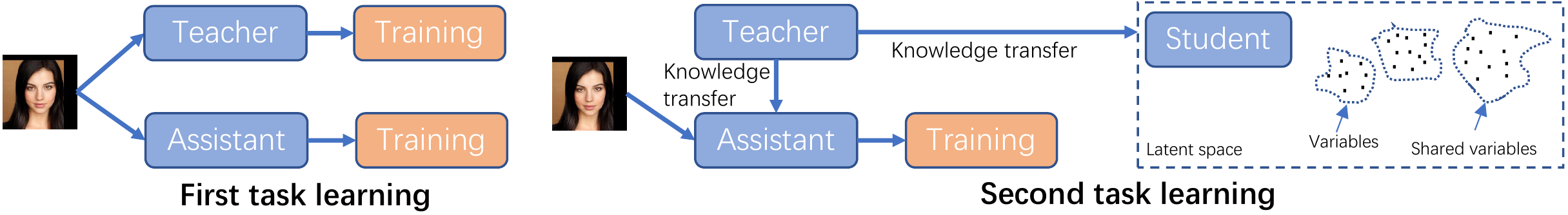}
	\centering
	\vspace{-10pt}
	\caption{Diagram showing the learning flow for LT-GANs. The Teacher and Assistant are trained jointly during the first task, while they exchange their roles, teaching each other in turns, when learning the following tasks.}
	\label{LTGANs_structure}
	\vspace{-15pt}
\end{figure*}

\vspace*{-0.5cm}
\section{Related works}
\vspace*{-0.2cm}

There are two categories of lifelong learning approaches~: memory-based systems \cite{AGEM}, and  generative modelling \cite{LifelongUnsupervisedVAE}. The former category of models uses a small buffer to store some data samples for each task which are then used to minimize the negative backward transfer when updating the network's parameters when learning a new task \cite{TinyLifelong}. However, such approaches require a significant computation processing \cite{AGEM}, when the number of tasks to be learnt increases. The generative modelling approaches would usually use a generator such as a Generative Adversarial Network (GAN) \cite{GAN} or a Variational Autoencoder (VAE) \cite{VAE} to reproduce previously learnt data samples before learning the next task. These generated data samples are then mixed with samples drawn from the current database, to form a new training data set for the model. Moreover, existing GAN based lifelong approaches \cite{LifelongInterpretable} can not learn inference models, which prevent their usefulness for a wide range of applications. In contrast, VAE based lifelong approaches \cite{LatentVAEGAN} are able to capture cross-domain representations over several tasks but lead to poor performance when learning databases of high complexity, given that VAEs used as generative replay networks tend to produce rather blurred images.

Another category of related works is based on the coupled \cite{CoupledGAN} or dual generative models \cite{dualGAN}. CoGANs \cite{CoupledGAN} consists of a pair of GANs which share their parameters for generator and discriminator networks. DualGAN \cite{dualGAN} has a similar network architecture with CoGAN, with the difference that DualGAN aims to learn an image-to-image translation framework while CoGAN aims to learn a joint distribution without accessing a tuple of corresponding images. The Twin Auxiliary Classiﬁers GANs (TAC-GAN) \cite{TwinAuxGANs} introduces a classifier as a new player to interact with the GAN's generator and discriminator in order to enforce the diversity in the generated data. In this paper we propose the Lifelong Adversarial Knowledge Distillation (LAKD) as the training algorithm which enables the LT-GANs model to learn new tasks without forgetting.

\vspace*{-0.2cm}
\section{Lifelong Twin Generative \\
Adversarial Networks (LT-GANs)}
\vspace*{-0.2cm}

Most lifelong learning models are applied in the context of supervised learning. The proposed Lifelong Twin Generative Adversarial Network (LT-GANs) consists of three components: two identical generators and a discriminator network. One of the generators is named the Teacher while the other one is the Assistant. LT-GANs training procedure is shown in the diagram from Fig.~\ref{LTGANs_structure}. Let ${\bf z}$ represent a random noise vector sampled from a Normal distribution $\mathcal{N}(0,{\bf I})$. The two generators, Teacher and Assistant, are parameterized by two identical neural networks $G_{\theta_T}({\bf z})$ and $G_{\theta_A}({\bf z})$, which aim to generates images ${\bf x}'_t$ and ${\bf x}'_a$ by taking the random noise vector ${\bf z}$ as input. The learning goal of LT-GANs in the first task is similar to that in GAN \cite{GAN}, by minimizing the distance between the data distribution and the distribution of the generated data. In this paper, we consider minimizing the Wasserstein distance, \cite{ImprWGAN}~: 
\begin{equation}
\vspace{-5pt}
\begin{aligned}
& \mathop {\min }\limits_{{G_{{\theta _T}}},{G_{{\theta _A}}}} \mathop {\max }\limits_{D \in \Theta} \underbrace{\mathbb{E}_{{\bf x}^1 \sim p({\bf x}^1)} [ D ({\bf x}^1) ] - \mathbb{E}_{{\bf x}'_{t} \sim p({{\bf{x}}_T})}  [D ( {\bf x}'_{t}) ]}_{\text{Teacher optimization}} \\&+ \underbrace{\mathbb{E}_{{\bf x}^1 \sim p({\bf x}^1)} [ D ({\bf x}^1) ] - \mathbb{E}_{{\bf x}'_{a} \sim p({{\bf{x}}_A})}  D ( {\bf x}'_{a}) ]}_{\text{Assistant optimization}} 
\end{aligned}
\end{equation}
where $\Theta$ represents a set of 1-Lipschitz functions. $p({\bf x}_T)$ and $p({\bf x}_A)$ represent the generator distributions for the Teacher and Assistant networks, $G_{\theta_T}({\bf z})$ and $G_{\theta_A}({\bf z})$, respectively, while $D(\cdot)$ represents the discriminator network. We introduce a gradient penalty term (momentum) \cite{ImprWGAN}, defined by $\lambda$, in order to enforce the Lipschitz constraint, resulting in:
\begin{equation}
\begin{aligned}
&\mathop {\min }\limits_{{G_{{\theta _T}}},{G_{{\theta _A}}}} \mathop {\max }\limits_{D \in \Theta} \mathbb{E}_{{\bf x}^1 \sim p({\bf x}^1)} [ D ({\bf x}^1) ] - \mathbb{E}_{{\bf x}'_{t} \sim p({{\bf{x}}_T})}  [D ( {\bf x}'_{t}) ] \\&+ \lambda \mathbb{E}_{\tilde{\bf x}_t \sim \mathbb{P}_{\tilde{\bf x}_T}} [ ( \left\| {\nabla _{\tilde{\bf x}_t}}
D ( \tilde{\bf x}_t) \right\|_2 - 1 )^2 ] \\
& + \mathbb{E}_{{\bf x}^1 \sim p({\bf x}^1)} [ D ({\bf x}^1) ] - \mathbb{E}_{{\bf x}'_{a} \sim p({{\bf{x}}_A})}  D ( {\bf x}'_{a}) ] \\ 
& +\lambda \mathbb{E}_{\tilde{\bf x}_a \sim \mathbb{P}_{\tilde{\bf x}_A}} [ ( \left\| {\nabla _{\tilde{\bf x}_a}}
D ( \tilde{\bf x}_a) \right\|_2 - 1 )^2 ].
\end{aligned}
\end{equation}
$\mathbb{P}_{\tilde{\bf x}_T}$ and $\mathbb{P}_{\tilde{\bf x}_A}$ are defined by sampling uniformly along straight lines between pairs of data from the given distribution $p({\bf x}^1)$, and the distribution generated by the Teacher net, and those sampled from $p({\bf x}^1)$ and the Assistant's distribution.

Traditional knowledge distribution approaches normally train a classifier on the predictions made by another classifier \cite{TowardKnowlege}. Some recent studies have proposed to learn a single model from an ensemble of networks in order to achieve a higher performance while requiring a lighter computational cost. These approaches, however, would require real data samples as well as supervision signals drawn from a single domain for knowledge distillation, which is a serious challenge for lifelong learning. In this paper, we propose the Lifelong Adversarial Knowledge Distillation (LAKD) for training LT-GANs. The main idea of LAKD is to encourage one of the generators to be a Teacher and to transfer its knowledge to another generator, which is the Assistant, during the lifelong learning. Let us assume that the Teacher and Assistant have been trained on the first task. When learning the second task, the Teacher has its parameters fixed and is seen as the knowledgeable source, while the Assistant learns data samples drawn from the Teacher's distribution as well as the given real data. During the third task learning, the Teacher and Assistant exchange their roles and the Assistant becomes now the Teacher and has its weights fixed, while transferring knowledge to the former Teacher which becomes now the Assistant. The LAKD loss is defined as:
\vspace*{-3pt}
\begin{equation}
\begin{aligned}
&\mathop {\min }\limits_G \mathop {\max }\limits_{D \in \Theta } {\mathbb{E}_{{\bf{x}} \sim p({{\bf{x}}^k})p({\bf{x}}_T^{k - 1})}}[D({\bf{x}})] - \\& {\mathbb{E}_{{\bf{x'}} \sim p({\bf{x}}_A^k)}}[D({\bf{x'}})]
 + \lambda \mathbb{E}_{\tilde{\bf x} \sim \mathbb{P}_{\tilde{\bf x}_A}} [ ( \left\| {\nabla _{\tilde{\bf x}}}
D ( \tilde{\bf x}) \right\|_2 - 1 )^2 ]
\end{aligned}
\vspace*{-3pt}
\end{equation}
where ${\bf x}$ are uniformly sampled from $p({\bf x}^k)$ and $p({\bf x}_T^{k-1})$, where $p({\bf x}_T^{k - 1})$ denotes the distribution $G_{\theta_T^{k-1}}({\bf z})$, generated by the Teacher, which had been trained on the $(k-1)$-th task. The fake data sample ${\bf x'}$ is drawn from $p({\bf x}_A^k)$ which represents the distribution $G_{\theta_A^k}({\bf z})$ generated by the Assistant following its training on the $k$-th task. Training LT-GANs using the proposed LAKD has many advantages when compared to other lifelong learning approaches \cite{LifelongUnsupervisedVAE,TinyLifelong}. Firstly, LAKD does not require to load previously learnt data samples \cite{MemoryReplayGAN} while its memory size does not change as the number of tasks increases. Secondly, it does not require to preserve the model's parameters or even a snapshot of these after each task switch, as we have in other generative replay methods \cite{LatentVAEGAN}.

\begin{figure*}[h]
\vspace{-5pt}
	\centering
		\includegraphics[scale=0.6]{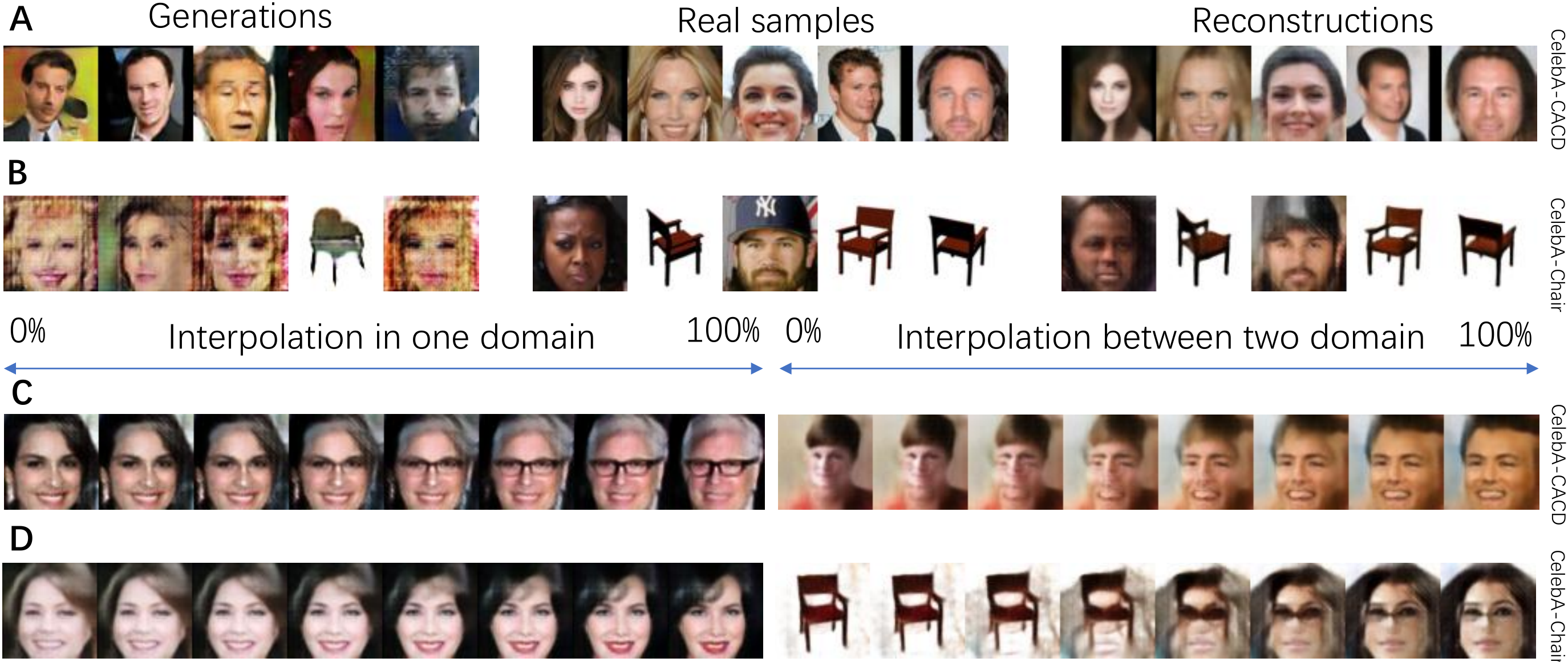}
	\vspace*{-0.2cm}
	\centering
	\caption{Image generation results when employing LT-GANs for the lifelong learning of CelebA-to-CACD and CelebA-to-3D Chairs databases are provided in A) and B). Interpolations in the latent space for the same databases are provided in C) and D).}
	\label{CelebaToChair_reco}
	\vspace*{-0.2cm}
\end{figure*}

\vspace*{-10pt}
\section{The Lifelong Teacher-Student}
\vspace*{-5pt}

We consider a Teacher-Student architecture, where the Teacher is represented by the LT-GANs, while the Student is implemented by a latent variable generative model $p({\bf x},{\bf z}) = p({\bf x}|{\bf z})p({\bf z})$. The marginal likelihood of $p({\bf x},{\bf z})$ is intractable given that it requires integration over the entire latent variable space $p({\bf x})=\int p({\bf x}|{\bf z})  p({\bf z})d{\bf z}$. Instead, we maximize the evidence lower bound (ELBO) on the sample log-likelihood, as in the Variational Autoencoder (VAE) inference, \cite{VAE}~:  
\begin{equation}
\begin{aligned}
\vspace*{-10pt}
 \log p ({\bf x}) & \ge  {\mathbb{E}_{{\bf z} \sim {q_{\varepsilon} } ({\bf z}|{\bf x})}[ \log p_{\omega} ({\bf x}|{\bf z})}] - D_{KL} [ q_{\varepsilon} ({\bf z}|{\bf x} )||p ({\bf z}) ] \\&={\mathcal L}_{\rm{VAE}} (\omega ,\varepsilon ) 
\label{DefineL_VAE}
\end{aligned}
\vspace*{-2pt}
\end{equation}
where $p_{\omega} ({\bf x}|{\bf z})$ is the probability implemented by the decoder and $q_{\varepsilon} ({\bf z}|{\bf x} )$ is that of the inference model, implemented by a neural network which has Gaussian-speciﬁc prior parameters $\{ \mu ,\sigma \} $ for its last layer's outputs, while $D_{KL}$ is the Kullback-Leibler (KL) divergence. The latent vector ${\bf z}$ is sampled using the reparametrisation trick ${\bf{z}} = {\bf \mu} + {\bf \gamma} \otimes {\bf \sigma}$, where ${\bf \gamma}$ is random noise drawn from $\mathcal{N}(0,{\bf I})$.

For learning cross-domain representations under the lifelong learning, we consider transferring the knowledge from the most knowledgeable generator to the Student network:
\begin{equation}
\vspace{-5pt}
\begin{aligned}
&\log [p({\bf x}^k)p({\bf x}_Q^{k - 1})] \ge \\&\underbrace{{\mathbb{E}_{{\bf{z}} \sim {q_\varepsilon }({\bf{z}}|{{\bf{x}}^k})}}[\log {p_\omega }({\bf{x}}|{\bf{z}})] - {D_{KL}}[{q_\varepsilon }({\bf{z}}|{{\bf{x}}^k})||p({\bf{z}})]}_{\text{Loss on data from k-th task}} +
 \\&
 \underbrace{{\mathbb{E}_{{\bf{z}} \sim {q_\varepsilon }({\bf{z}}|{\bf{x'}})}}[\log {p_\omega }({\bf{x}}|{\bf{z}})] - {D_{KL}}[{q_\varepsilon }({\bf{z}}|{\bf{x'}})||p({\bf{z}})] }_{\text{Knowledge distillation loss}}= {\mathcal{L}_{stu}}(\omega,\varepsilon )
\label{stuLoss}
\end{aligned}
\vspace{5pt}
\end{equation}
where $p({\bf x}_Q^{k - 1})$ can be either $p({\bf x}_A^{k - 1})$ or $p({\bf x}_T^{k - 1})$, depending on which network is more knowledgeable when learning the $k$-th task. ${\bf x}'$ is sampled from $p({\bf x}_Q^{k-1})$. The Student network training is synchronised with that of LT-GANs.

We enable the Student for learning dissentangled representations \cite{DeepVAEMixture,JointVAE_MI} by penalizing the Kullback-Leibler divergence between the posterior and prior distributions, \cite{baeVAE}~:
\begin{equation}
\begin{aligned}
\log [p({\bf x}^k)p({\bf x}_Q^{k - 1})] &\ge {\mathbb{E}_{{\bf{z}} \sim {q_\varepsilon }({\bf{z}}|{{\bf{x}}^k})}}[\log {p_\omega }({\bf{x}}|{\bf{z}})] \\&- \beta {D_{KL}}[{q_\varepsilon }({\bf{z}}|{{\bf{x}}^k})||p({\bf{z}})] \\&+ 
{\mathbb{E}_{{\bf{z}} \sim {q_\varepsilon }({\bf{z}}|{\bf{x'}})}}[\log {p_\omega }({\bf{x}}|{\bf{z}})] \\&- \beta {D_{KL}}[{q_\varepsilon }({\bf{z}}|{\bf{x'}})||p({\bf{z}})] = \mathcal{L}_{Dis}(\omega,\varepsilon )
\label{Disentangled_Loss}
\end{aligned}
\end{equation}
where $\beta = 1$ corresponds to (\ref{stuLoss}). A large $\beta$ encourages the independence between latent variables but sacrifices the reconstruction quality. During the experiments we set $\beta = 4$.

\begin{figure}[h]
\hspace*{-6pt}
\centering
	\subfigure[Changing facial expression.]{
		\centering
		\includegraphics[scale=0.49]{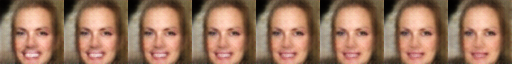}
	}
	\hspace*{-6pt}
	\subfigure[Changing the shape of a chair.]{
		\centering
		\includegraphics[scale=0.49]{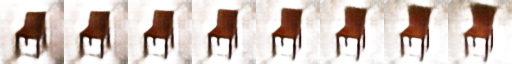}
	}
	\vspace*{-10pt}
	\caption{Disentangled representation results after the lifelong learning of CelebA to 3D-Chairs databases. }
	\label{Disentangled}
	\vspace{-7pt}
\end{figure}

\begin{figure}[]
\vspace{-6pt}
\hspace*{-0.55cm}
	\centering
	\subfigure[NLOG on various datasets.]{
		\centering
	\includegraphics[scale=0.76]{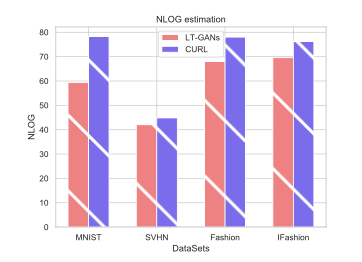}
	} \hspace*{-0.95cm}
	\subfigure[NLOG and MSE on MNIST.]{
		\centering
			\includegraphics[scale=0.76]{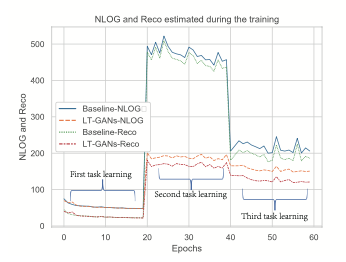}
	}
	\centering
	\vspace{-10pt}
	\caption{Analysis results for LT-GANs.}
	\label{avgRecoAcrossModels}
	\vspace{-12pt}
\end{figure}

\vspace*{-0.3cm}
\section{Experiments}
\vspace*{-0.2cm}

We train the proposed model, using the ELBO criterion, (\ref{stuLoss}). The results generated after the lifelong learning one database under the CelebA \cite{Celeba} 
to CACD \cite{CACD}, and CelebA to 3D-Chair database lifelong learning, shown in Fig.~\ref{CelebaToChair_reco}A and \ref{CelebaToChair_reco}B, respectively. Then we interpolate between two latent vectors encoding two different images, from the same database and also from different databases, with the results shown in Fig.~\ref{CelebaToChair_reco}C and \ref{CelebaToChair_reco}D, for the same databases as above. It observes that an image can be smoothly transformed into another, even when the two images come from two different domains, as when interpolating between a 3D chair and a face image, as shown in the second example from Fig.~\ref{CelebaToChair_reco}D. These results indicate that the Student can capture shared and domain-specific generative factors over time.

\begin{table}[h]
    \centering
    \small
    \begin{tabular}{lccc}
    \toprule
		\cmidrule(r){1-4}
		  Dataset & LT-GANs & CURL \cite{LifelongUnsupervisedVAE}  & LGM  \cite{GenerativeLifelong}   \\
		\hline
		MNIST &\textBF{878.92} & 887.40 & 900.18  \\
		SVHN &\textBF{219.15} &261.08&262.20  \\
		Fashion &\textBF{365.62} &639.19&642.20 \\
		Omniglot &\textBF{514.87} &695.68&700.84 \\
		\textBF{Average} &\textBF{494.64} &620.83&626.35 \\
			\bottomrule
	\end{tabular}
	\vspace{-2pt}
		\caption{The negative log-likelihood estimation for the lifelong learning of MNIST, SVHN, Fashion and Omniglot.}
	\label{LogLikelihood_longTask}
	\vspace{-4pt}
\end{table}

\begin{table}[h]
    \centering
    \vspace{-10pt}
    \begin{tabular}{lcc}
		\toprule
		\cmidrule(r){1-3}
		Tasks & LT-GANs & CURL \cite{LifelongUnsupervisedVAE}    \\
		\midrule
		First task & \textBF{62.85} &155.59   \\
		Second task & \textBF{59.27}&166.47   \\
		Third task & \textBF{60.35}&169.28   \\
		\bottomrule
	\end{tabular}
	\vspace{-2pt}
		\caption{FID score after the lifelong learning of CIFAR10, CIFAR100, Sub1- and Sub2-ImageNet databases.}
	\label{ISTab2}
	\vspace{-8pt}
\end{table}

We evaluate the performance of various lifelong learning models when training LT-GANs using MNIST, SVHN, Fashion and Omniglot databases (MSFO sequence). The negative log-likelihood (NLOG) results, estimated as the reconstruction error plus the KL term, are provided in Table~\ref{LogLikelihood_longTask}. LT-GANs performs better in all tasks, while VAE based lifelong methods, such as CURL \cite{LifelongUnsupervisedVAE} and LGM \cite{GenerativeLifelong} tend to forget previously learnt tasks. We train various models under the lifelong learning of CIFAR10 and other two distinct datasets, called Sub1 and Sub2, which are subsets of the ImageNet database \cite{ImageNet_1}. In Table~\ref{ISTab2} we evaluate the quality of the generated images by calculating the Fréchet Inception Distance (FID) \cite{FID}, after each task switch during the lifelong learning.

\begin{table}[h]
\vspace{-2pt}
    \centering
    \begin{tabular}{lccc}
    \toprule
		\cmidrule(r){1-4}
		 Dataset & LT-GANs & CURL \cite{LifelongUnsupervisedVAE}  & LGM  \cite{GenerativeLifelong}   \\
		\hline
		MNIST-c & \textBF{139.98}& 437.45 & 1365.41 \\
		SVHN-c & \textBF{141.57}& 209.70 & 206.44  \\
		Fashion-c &\textBF{51.68} & 54.72 & 145.78   \\
		\textBF{Average} & \textBF{111.08}& 233.96 & 556.13  \\
			\bottomrule
	\end{tabular}
	\vspace{-2pt}
		\caption{NLOG for the lifelong learning with poorly defined task boundaries on MNIST-c, SVHN-c and Fashion-c.}
	\label{LogLikelihood_taskBounders}
	\vspace{-9pt}
\end{table}

We also train the proposed Teacher-Student framework under the CelebA to 3D-Chairs by using the loss defined by (\ref{Disentangled_Loss}). We then manipulate the latent space by changing a single latent variable while fixing the others. The  generated image results when using the LT-GANs model are shown in Figs.~\ref{Disentangled}-a and \ref{Disentangled}-b, when changing the facial expression of a woman, and the shape of a chair from CelebA and 3D chair databases.

In order to test the robustness of the lifelong generative models we consider fuzzy task boundaries. In these experiments we exchange certain images, from one of the classes, between two databases and
create new databases, MNIST-c, SVHN-c and Fashion-c, while preserving the images for the other nine classes. The NLOG results are provided in Table~\ref{LogLikelihood_taskBounders}, where the proposed Teacher-Student framework still achieves the best results in terms of NLOG image reconstruction.

We also split the database into data sets containing images of a certain class, corresponding to 10 tasks in total. We train various models on one database with 10 tasks and evaluate NLOG on all testing samples, and the results are provided in Fig.~\ref{avgRecoAcrossModels}-a. From this bar-plot it can be observed that the proposed framework outperforms CURL \cite{LifelongUnsupervisedVAE}. We investigate the effect of the proposed LAKD loss function by changing $\beta$ in (\ref{Disentangled_Loss}) for the lifelong learning in terms of the log-likelihood and reconstruction errors measured as the MSE error. LAKD loss function plays an important role in overcoming forgetting, according to Fig.~\ref{avgRecoAcrossModels}-b, where the baseline is considered when the Teacher and Assistant are trained jointly in each task. 

\vspace*{-0.15cm}
\section{Conclusion}
\vspace*{-0.1cm}

We introduce a new lifelong learning LT-GANs model, made up of a dual-generator network, which is trained in a memory-efficient and end-to-end learning manner using the proposed LAKD loss function. We further extend the LT-GANs model into a Teacher-Student framework in order to capture data representations, where the two generators teach alternatively one another, as well as to a Student network. The proposed framework is enabled with the ability to model disentangled representations under the unsupervised lifelong learning setting. It is also shown to generate smooth interpolations between images associated with different databases. The results demonstrate that the proposed framework achieves the state of the art in lifelong unsupervised representation learning.

\bibliographystyle{IEEEbib}
\bibliography{VAEGAN.bib}

\end{document}